\documentclass[11pt,a4paper]{article}
\usepackage[margin=1in]{geometry}
\usepackage{setspace}
\usepackage{parskip}
\usepackage{hyperref}
\usepackage{enumitem}
\usepackage{times}
\usepackage{authblk}
\usepackage{graphicx}
\usepackage{amsmath}
\usepackage[style=numeric-comp, sorting=none, doi=true, url=false, isbn=true, eprint=true, maxbibnames=99, backend=biber]{biblatex}
\renewbibmacro{in:}{}
\usepackage{comment}
\usepackage{xcolor}

\title{\vspace{-50 pt}\textbf{Probabilistic Algorithms for Ising Machines
from Optimization to Generative AI}}
\author[1,*]{Corentin Delacour}
\author[1]{Xiuqi Zhang}
\author[1]{Abdelrahman S. Abdelrahman}
\author [1,2]{Saleh Bunaiyan}
\author[1]{Kyle Lee}
\author[1]{Shuvro Chowdhury}
\author[1]{Kerem Y. Camsari}
\affil[1]{Department of Electrical and Computer Engineering, University of California, Santa Barbara, Santa Barbara, CA 93106, USA}
\affil[2]{Electrical Engineering Department, King Fahd University of Petroleum $\&$ Minerals (KFUPM), Dhahran 31261, Saudi Arabia}
\affil[*]{Corresponding author: Corentin Delacour  (\href{mailto:delacour@ucsb.edu}{delacour@ucsb.edu})}

\date{}

\begin{document}
\maketitle
\vspace{-10 pt}

\section*{Abstract}
Ising machines have emerged as promising hardware accelerators for
intractable optimization and sampling problems, yet their practical
impact increasingly hinges on the co-design of algorithms and
hardware, where algorithmic demands shape new architectures and new
hardware capabilities inspire entirely new algorithms. In this Review,
we survey probabilistic algorithms designed for portability across
diverse Ising platforms, advocating a top-down perspective that
prioritizes principled methods with provable guarantees. We cover
foundational methods such as simulated annealing and parallel
tempering, including two-dimensional extensions that natively encode
hard constraints, and examine approaches that expand the scale of
solvable problems from cluster mean-field methods to variational
samplers. We highlight the Probabilistic Approximate Optimization
Algorithm (PAOA), a classical analog of QAOA that emerged directly
from probabilistic hardware development, and explore how generative
AI and Ising machines might reinforce each other: learned models propose
global moves to accelerate optimization, while probabilistic
techniques improve inference in large language models. Much as quantum
computing has seen algorithms co-evolve with hardware, probabilistic and Ising
computing stand at a similar inflection point. We outline a co-design
framework for accelerating the capabilities and adoption of
next-generation Ising machines.


\begin{figure}[t!]
    \hspace{-30 pt}
    \includegraphics[width=1.05\linewidth]{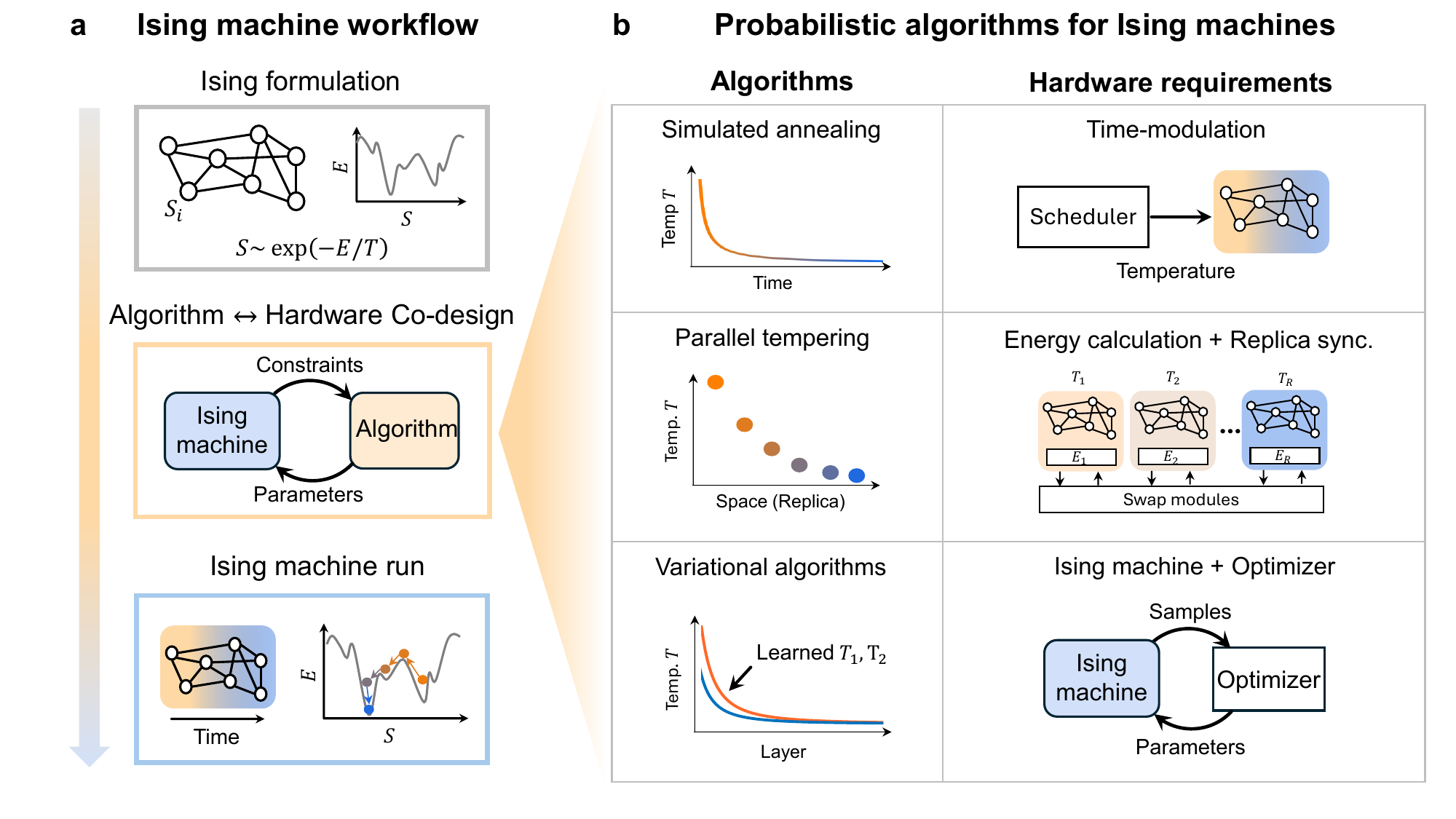}
   \caption{\textbf{Ising machine workflow and probabilistic 
algorithm-hardware co-design.} (a) Top: an optimization or 
sampling task is encoded as an Ising problem with energy 
$E$ and Boltzmann distribution $S \sim \exp(-E/T)$ with $T$ the temperature parameter. Middle: 
the algorithm and Ising machine co-design loop, where 
algorithmic demands set hardware parameters and hardware 
constraints shape algorithm choice. Bottom: the Ising 
machine explores the energy landscape over time and returns 
samples $S$. (b) Three families of probabilistic algorithms 
surveyed in this Review, paired with their hardware 
requirements. Simulated annealing requires time-modulation 
of temperature via a scheduler. Parallel tempering fixes 
distinct temperatures across replicas in space and requires 
energy calculation and replica synchronization for state 
swaps. Variational approaches couple an Ising machine with a classical optimizer that iteratively updates parameters based on sampled states. The Probabilistic Approximate Optimization Algorithm (PAOA) showcases this by learning temperatures across layers.}
    \label{fig1}
\end{figure}

\section{Introduction}

Ising machines are specialized hardware accelerators designed to tackle intractable combinatorial optimization and sampling tasks, originally inspired by the physics of interacting spins \cite{Mohseni2022}. In its simplest form, a network of binary spins $S_i=\pm1$ is governed by the pairwise Ising Hamiltonian:
\begin{equation}
    E=-\sum_{i<j}J_{ij}S_iS_j-\sum_ih_iS_i 
    \label{eq_Ising}
\end{equation}
where $J_{ij}=J_{ji}$ represents symmetric coupling weights between spins, and $h_i$ the external biases applied to individual nodes. Finding the ground state of this discrete Hamiltonian is famously NP-hard, and equivalent to a wide range of combinatorial optimization problems by appropriately configuring the coefficients $J_{ij}$ and $h_i$ \cite{lucas2014ising}. However, many problems of practical interest push beyond this canonical quadratic, binary, symmetric formulation in at least three directions. First, higher-order interactions lead to polynomial unconstrained binary optimization (PUBO) formulations that generalize Eq.~\ref{eq_Ising} to $k$-body tensors \cite{bybee2023,nikhar2024,bhattacharya2024,hutchinson2026}. Second, variables with more than two states motivate Potts-model and encoded-variable extensions \cite{duffee_2025,callahancoray2026restoringsparsitypottsmachines,iyer2025,Garg2025}. Third, relaxing the symmetry constraint ($J_{ij}\neq J_{ji}$) yields directed graphical models such as Bayesian and feedforward stochastic networks, for which no closed-form energy function exists and sequenced updating must be enforced \cite{nihal_IEDM23,Hong2023,ghantasala2025improvingdeepneuralnetwork}. These generalizations are each the subject of active algorithmic and hardware development, and the probabilistic methods discussed in this Review apply broadly to such extended formulations. The fundamental motivation behind Ising machines is to natively instantiate these computational models in physical hardware, thereby exploiting massive parallelism and ultrafast intrinsic dynamics to solve the hard problems encoded within their parameters (Figure~\ref{fig1}a).

Historically, the development of Ising machines has been largely driven by a physicist's perspective, leveraging the natural dynamics of classical or quantum systems to minimize an energy functional corresponding to the Ising Hamiltonian \cite{Mohseni2022}. A non-exhaustive list of these physics-based approaches includes coherent Ising machines \cite{yamamoto_2013,inagaki_coherent_2016,mcmahon2016fully,honjo_2021}, photonic Ising machines \cite{Pierangeli2019,RoquesCarmes2020,Gao2024,Kalinin2025,Veraldi2025,aadhi2026photonic}, oscillatory Ising machines \cite{wang2019_oim,bashar_experimental_2021,ahmed_probabilistic_2021,Maher2024,Graber2024,Yuqi2024,Chen2024,lianlong2025,Behera2026,Ekanayake2026}, and memristor-based annealers \cite{Fahimi2021,Jiang2023,Yue2024,hutin_2025}. In parallel, a complementary class of digital Ising machines has emerged, executing physics-inspired dynamics on programmable CMOS substrates: examples include Fujitsu's Digital Annealer \cite{aramon2019physics}, Hitachi's CMOS annealing accelerator \cite{hitachi2016ising}, simulated bifurcation on FPGAs and GPUs \cite{simulated_bifurcation_2019,goto_high-performance_2021,tatsumura_scaling_2021,Kanao2022}, and FPGA-based probabilistic computers \cite{aadit_2022,chowdhury2025pushing}. For a comprehensive overview of these physical and digital implementations, we direct the reader to the review by Mohseni \textit{et al.} \cite{Mohseni2022}.

While Ising machines have demonstrated remarkable potential, sometimes scaling to hundreds of thousands of nodes \cite{honjo_2021}, many physics-based approaches rely on intrinsic physical dynamics that amount to sophisticated heuristics: they lack detailed balance, offer no formal convergence guarantees, and cannot sample from a prescribed probability distribution. This limits their utility beyond optimization, ruling out tasks such as Bayesian inference or generative model training that require controlled exploration of the full state space. By contrast, Markov Chain Monte Carlo (MCMC) methods provide these guarantees by construction: they satisfy detailed balance, converge to well-defined target distributions, and serve as the foundation for advanced strategies such as parallel tempering, cluster methods, and variational sampling. This motivates a top-down design question: rather than asking what algorithm a given device naturally executes, one asks how principled algorithms can be efficiently mapped onto, or co-designed with, specialized hardware.

A natural bridge between MCMC algorithms and Ising hardware emerges when the device dynamics themselves implement a principled sampling scheme. Programmable digital Ising machines, whether built from FPGAs \cite{aadit_2022,chowdhury2025pushing}, custom CMOS \cite{smithson2019efficient,aramon2019physics,hitachi2016ising}, or hybrid nanodevice platforms \cite{Singh2024,kemal_2025,whitehead2023cmos,borders2019integer,li2026ultralowpowerfastising,Yang2026}, can be configured to execute Gibbs or Metropolis updates, satisfying detailed balance by construction. The concept of \textit{probabilistic bits} (p-bits) \cite{camsari2017stochasticL}, whose physical realizations include nanosecond-scale stochastic magnetic tunnel junctions \cite{hayakawa2021nanosecond,safranski2021demonstration}, makes this correspondence especially transparent: a p-bit is a stochastic binary unit whose output probability is a sigmoidal function of its inputs, functioning as a tunable building block for Boltzmann machines \cite{ackley1985learning}. Coupling p-bits with symmetric weights and biases directly implements Gibbs sampling for the Boltzmann distribution $\pi\propto\exp(-\beta E)$, while relaxing symmetry or extending to continuous (g-bit \cite{nihal_gbit}) and multi-state (Potts \cite{kalinin2018simulating,whitehead2023cmos,duffee_2025}) variants broadens the class of representable models, as discussed earlier. Crucially, this top-down, algorithm-first design means that any advance in MCMC methodology, parallel tempering, cluster updates, variational schemes, can be immediately inherited by the hardware, a principle that organizes the remainder of this Review. For a thorough exploration of p-bit physics and device architectures, we refer to the review by Chowdhury \textit{et al.} \cite{chowdhury_review}. In our view, probabilistic computing in this sense is not defined by any single device technology, whether magnetic tunnel junctions, CMOS ring oscillators, or photonic noise sources, but rather by the principle that hardware and algorithms are co-designed around controlled, programmable randomness.

The goal of this Review is to survey probabilistic algorithms through a hardware-software co-design lens (Figure~\ref{fig1}b), showing how strategies developed in computational physics and machine learning can be adapted to a broad spectrum of Ising accelerators, from exact MCMC samplers to approximate and deterministic solvers and how hardware constraints in turn inspire new algorithmic ideas.  Indeed, this co-design is bidirectional: just as algorithmic
requirements have driven new hardware architectures, the availability
of dedicated probabilistic hardware has catalyzed fundamentally new algorithms such as PAOA \cite{Combes_2025,PAOA_OPUS}, which arguably would not have emerged from a purely software-oriented perspective. In this respect,
the trajectory of probabilistic computing mirrors that of quantum computing, where algorithms, applications, and hardware have
co-evolved in tandem
\cite{farhi2022qaoa,preskill2018quantum,mohseni2024build}.

We organize the discussion along a progression of increasing algorithmic sophistication: from foundational annealing methods and their hardware-centric implementations (Section~\ref{sec_annealing}), to parallel tempering and its recent two-dimensional extension for natively enforcing hard constraints (Section~\ref{sec_pt}). We then address the physical limits of single-chip systems through cluster mean-field algorithms that enable capacity scaling via hybrid CPU--accelerator architectures (Section~\ref{sec_cmft}), and variational approaches that autonomously discover problem-tailored heuristics (Section~\ref{sec_variational}). Finally, we highlight synergies between generative AI and Ising computation, where learned models propose global state updates to accelerate optimization and, conversely, probabilistic Ising-like primitives reduce memory traffic in AI inference (Section~\ref{sec_synergies_AI}).

\section{Annealing algorithms and their hardware realizations} \label{sec_annealing}

Simulated annealing (SA) is a stochastic optimization algorithm inspired by the physical process of cooling materials to achieve low-energy, defect-free states \cite{kirkpatrick1983optimization}. The algorithm maps a cost function to a physical energy $E$ such as the Ising Hamiltonian (Eq.~\ref{eq_Ising}). The states $S$ are sampled according to the Boltzmann distribution:
\begin{equation}
    \pi =\frac{\exp\big(-\beta E\big) }{Z}
    \label{eq:boltzmann}
\end{equation}
where $Z$ represents the partition function, and $\beta$ is the inverse temperature. At low $\beta$ (high temperature), the system explores the energy landscape freely; as $\beta$ increases according to a cooling schedule, the search transitions from coarse-grained global exploration to fine-grained, energy-minimizing local moves.

Sampling from the Boltzmann distribution (Eq.~\ref{eq:boltzmann}) naturally maps to the dynamics of probabilistic bits (p-bits) \cite{camsari2017stochasticL}, where each p-bit produces a probabilistic spin update conditioned on its neighbors, equivalent to Gibbs sampling. While conceptually simple, SA imposes a time-dependent constraint on hardware: the ability to modulate system-wide fluctuations dynamically (Figure~\ref{fig1}b). Ideally, this modulation should be intrinsic and global, avoiding the need to address individual devices. In digital architectures, annealing often incurs an arithmetic overhead, requiring the on-chip multiplication of synaptic inputs by $\beta$ \cite{PAOA_OPUS} or the reloading of weights scaled by $\beta$ \cite{aadit_2022}. A key co-design goal is to minimize this overhead: if the time required to update the temperature exceeds the device's sampling time, the hardware loses the speed advantage of its intrinsic dynamics. Analog implementations address this by mapping $\beta$ to a single global physical variable: the operating voltage in coupling memristors \cite{Jiang2023,hutin_2025} or the tunable autocorrelation time in random number generators \cite{kemal_2025}.

The same time-domain annealing principle underlies a broader family of continuous-variable Ising platforms. In Coherent Ising Machines (CIMs) \cite{inagaki_coherent_2016, mcmahon2016fully, Yamamoto2017, honjo_2021}, gradually increasing the pump amplitude of optical parametric oscillators binarizes optical phases, so that the collective mode with the lowest effective loss, corresponding to the relaxed ground state, reaches threshold and oscillates first. Similarly, Oscillatory Ising Machines (OIMs) \cite{wang2019_oim, bashar_experimental_2021,Maher2024, Graber2024} minimize the continuous phase-domain energy (equivalent to the XY model) while annealing a harmonic injection strength to enforce phase binarization. Simulated Bifurcation (SB) \cite{simulated_bifurcation_2019, goto_high-performance_2021, tatsumura_scaling_2021} numerically models a classical Hamiltonian, introducing an inertial momentum term that enables chaotic exploration to escape local minima as the pumping amplitude increases. Across all these approaches, along with weight-annealing methods \cite{Fahimi2021,Jiang2023}, a global control parameter is smoothly modulated to traverse a continuous relaxation of the discrete energy landscape \cite{kadowaki_1998}. While these physics-driven dynamics are highly effective in practice, successfully scaling to 100,000 spins in parallel and time-multiplexed implementations \cite{honjo_2021,goto_high-performance_2021}, their relationship to equilibrium sampling remains an open question, and reaching the ground state lacks the theoretical convergence guarantees of MCMC-based annealing.

Simulated quantum annealing (SQA) occupies a distinct position. Through the Suzuki-Trotter transformation \cite{suzuki1976relationship}, a $d$-dimensional quantum system is mapped onto a $(d+1)$-dimensional classical system comprising coupled Trotter replicas \cite{camsari2019scalable,chowdhury2025pushing}. Because each replica is updated via Gibbs sampling, SQA inherits the convergence guarantees of MCMC while emulating quantum tunneling through inter-replica coupling \cite{denchev2016computational}. Benchmarking studies of quantum versus classical annealing on tunneling-dominated landscapes have provided much of the empirical motivation for these comparisons \cite{boixo2014evidence}. Hardware implementations have realized SQA on FPGA-based probabilistic computers \cite{grimaldi2022,Onizawa2026}, scaling to 32 replicas and 32,768 total spins \cite{SQA_hariyama_2021}. Performance comparisons indicate that the relative advantage of SQA over the straightforward approach of best-of-R SA is problem-dependent \cite{grimaldi2022,raimondo_2025}, motivating the more flexible exchange-based methods discussed next.

Importantly, SA's reliance on a single, hand-tuned cooling schedule raises a natural question: can the schedule itself or multiple schedules simultaneously be {learned} from the problem structure? The Probabilistic Approximate Optimization Algorithm (PAOA), discussed in Section~\ref{sec_variational}, answers this affirmatively: it recovers SA as a special case with a single schedule and extends it to multi-schedule, out-of-equilibrium trajectories that are variationally optimized for a given problem class \cite{PAOA_OPUS}.

\section{Parallel Tempering for Ising machines} \label{sec_pt}
\begin{figure}[t!]
    \hspace{-22pt}
    \includegraphics[width=1.12\linewidth]{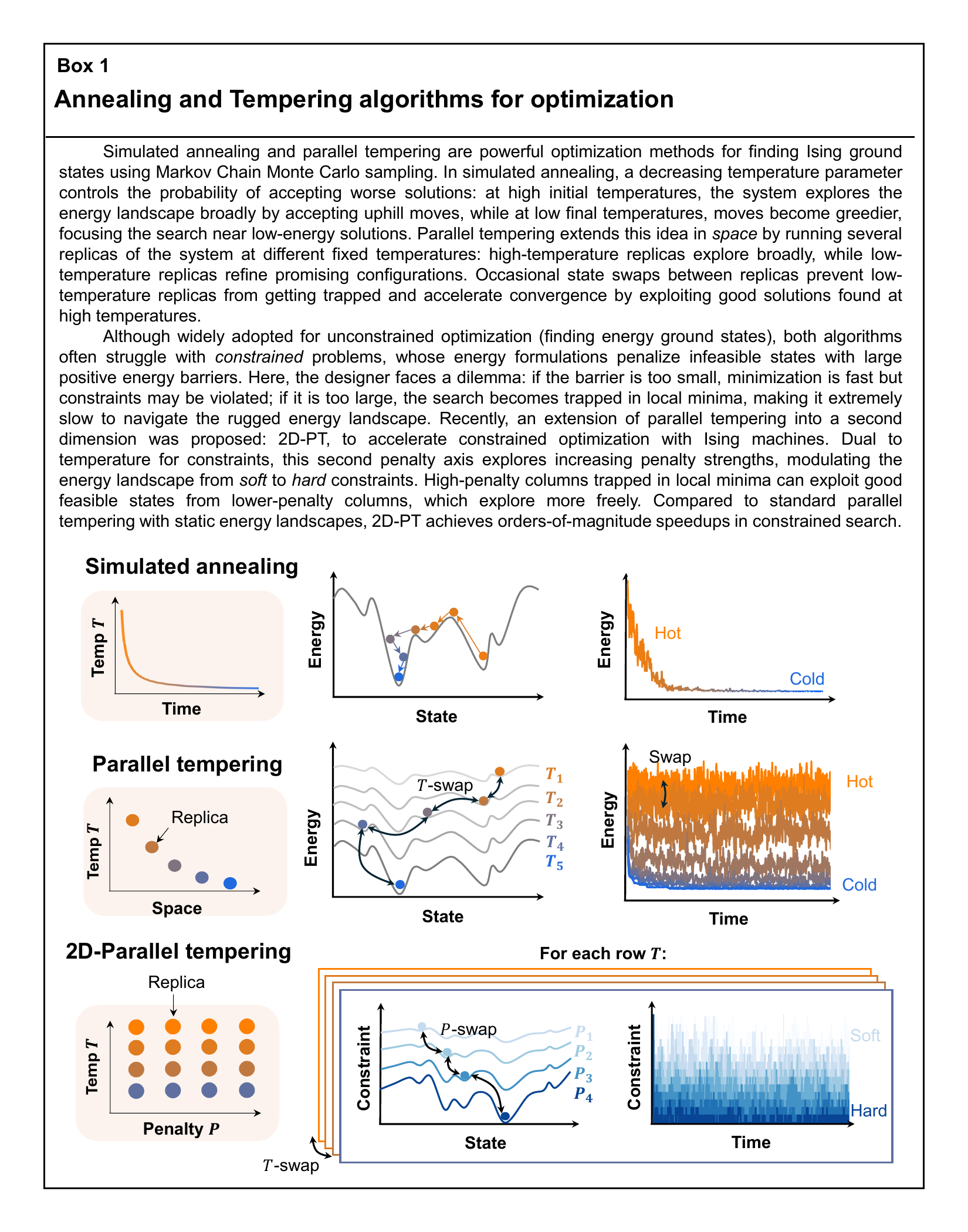}
    \label{Box1}
    \vspace{15pt}
\end{figure}
Instead of annealing the temperature or shaping the energy landscape in the \textit{time} domain,  exchange Monte Carlo \cite{Swendsen_1986, hukushima1996exchange}, simulated tempering \cite{marinari1992simulated}, or Parallel Tempering (PT) \cite{earl2005parallel} modulate the temperature in \textit{space} using multiple replicas of the system at different temperatures. The key idea is to exploit higher-temperature replicas, which are less likely to become trapped in local minima, and exchange their promising states with colder, greedier replicas that mostly accept energy-minimizing moves. These exchanges, or swaps, act as global moves in the energy landscape to boost MCMC exploration and find low-energy states more rapidly. Typically, two adjacent replicas in the temperature ladder swap their states according to the Metropolis probability:
\begin{equation}
    P_{\rm swap}=\min\big[1,\exp(\Delta \beta \Delta E) \big ]
    \label{swap}
\end{equation}
where $\Delta \beta $ and $\Delta E$ are the differences of inverse temperatures and energies between the two replicas. PT consists of an outer loop for swap moves (determined by Eq.~\ref{swap}), and an inner loop for Monte Carlo sampling within each replica in parallel. Compared to SA, which modulates temperature in time, each replica samples at its own temperature during the whole execution. This implies that PT can also be used as a sampling algorithm since the swap moves satisfy detailed balance.
SA and PT are further compared in Box~1. The inherent parallelism in PT is very suitable for Ising machines, which can physically map each replica to an Ising processor with a fixed temperature during the process.

\subsection{Parallel tempering in hardware}

A fundamental but underexplored question is whether PT's overhead, maintaining multiple replicas and synchronizing swaps, is justified compared to a simpler baseline: running $R$ independent SA instances and returning the best result ($R$-SA). Because $R$-SA requires no inter-replica communication, it is trivially parallelizable on any Ising hardware. Adaptive PT on FPGA-based p-computers has been shown to outperform both conventional PT and $R$-SA on spin-glass benchmarks \cite{aadit_2023}, but the margins can be modest, particularly for weakly frustrated instances. Given $R$-SA's simplicity, this comparison should inform any hardware investment in PT infrastructure: the swap mechanism must demonstrably earn its keep.

When PT is warranted, its hardware realization incurs two costs absent in SA: global energy calculation for each replica and synchronization for swap moves (Eq.~\ref{swap}). Unlike SA, where update decisions are strictly local, PT requires global state awareness. These overheads become critical bottlenecks in regimes requiring frequent communication, such as swapping at every sweep, found to be optimal for 3D spin glasses~\cite{chowdhury2025pushing}. Several architectures have explored different trade-offs in managing this cost. Fujitsu's Digital Annealer incorporates PT natively, with its third-generation architecture operating exclusively in PT mode for fully connected problems up to 100,000 variables \cite{nakayama2021description}. Fully digital FPGA implementations have demonstrated multi-spin updates and multi-replica swaps for Max-cut instances up to 1,024 nodes~\cite{Yao_2024}, while larger 2,700-spin 3D spin glass instances trade strict spatial parallelism for time-multiplexing, scaling to 33 temperature replicas each augmented by 4 Isoenergetic Cluster Move replicas (132 total)~\cite{chowdhury2025pushing,zhu_efficient_2015}. Other recent optimizations include accelerating PT with in-memory computing primitives combined with WalkSAT heuristics~\cite{beausoleil_2025} and sparsifying all-to-all graphs for color-based parallel updates in multiple-input multiple-output (MIMO) decoding~\cite{sajeeb_2026}. A complementary approach keeps dense connectivity but introduces an inertia term in the p-bit dynamics, enabling fully parallel synchronous updates with time-to-solution gains on Max-Cut, Sherrington-Kirkpatrick, and 5G MIMO benchmarks \cite{zhu2026fully}.

\subsection{Enforcing constraints with 2D-Parallel Tempering}

Ising machines excel at unconstrained energy minimization, but real-world applications typically demand the satisfaction of strict constraints. A classic example is the Traveling Salesman Problem (TSP), where mapping an $N$-city tour to an Ising model requires $N^2$ spins with one-hot constraints enforcing that each city appears once and each position is unique~\cite{hopfield_neural_1985}. Standard approaches encode these requirements using the penalty method, which adds energy terms to prevent invalid states. However, the penalty strength introduces a critical trade-off: too weak and the ground state may violate constraints; too strong and steep energy barriers trap the system in local minima~\cite{sajeeb_2025}. Lagrange relaxations that dynamically tune penalties in an outer loop can mitigate this, showing orders-of-magnitude reductions in the Monte Carlo budget for quadratic knapsack problems~\cite{delacour_SAIM}, but they require reprogramming the Ising machine's weights at every outer iteration, interrupting the sampling pipeline.

2D-Parallel Tempering (2D-PT) addresses this dilemma by parallelizing the search across both energy and feasibility landscapes~\cite{lee2015enhancing,2DPT}. Where classical PT interpolates across temperatures, 2D-PT adds an orthogonal axis that interpolates across penalty strengths, forming a two-dimensional array of replicas (Box~1). Each column acts as a classical PT instance at a fixed penalty; lower-penalty columns are less constrained, allowing them to discover feasible states that are subsequently transferred to higher-penalty replicas via a new swap move with probability:
\begin{equation}
    P_{\rm{swap-P}}=\min\big[1,\exp(\beta\Delta P\Delta g)]
\end{equation}
where $\beta$ is the inverse temperature of the corresponding row, $\Delta P$ the penalty difference between columns, and $\Delta g$ the constraint violation difference ($g=0$ when constraints are satisfied). The $\beta$ prefactor makes lower rows greedier, accepting only favorable swaps.

This mechanism is particularly effective for a pervasive hardware challenge: copy constraints. Because physical Ising machines often possess fixed, limited connectivity, embedding dense problems requires mapping a single logical node to multiple physical spins. These copies severely disrupt the ground state search, even with finely tuned penalty strengths~\cite{sajeeb_2025}. 2D-PT with adaptive scheduling circumvents manual tuning entirely, translating to massive performance gains: orders-of-magnitude speedups over classical PT for graph sparsification on Wishart instances~\cite{2DPT}, a $250\times$ acceleration for 64-node Sherrington-Kirkpatrick spin glasses, and over $10\times$ decoding speedup for 128-node MIMO problems~\cite{sajeeb_2026}. Recently, the same principle of decoupling constraint and objective temperatures has been applied to TSP, where fixing the constraint temperature at a pre-tuned value and tempering only the target energy reduces the problem to a standard one-dimensional PT ladder with improved convergence on traveling-salesman instances \cite{grimaldi2026}.

\section{Scaling Ising Machines beyond single-chip limits} \label{sec_cmft}

\begin{figure}[t!]
    \centering
    \includegraphics[width=\linewidth]{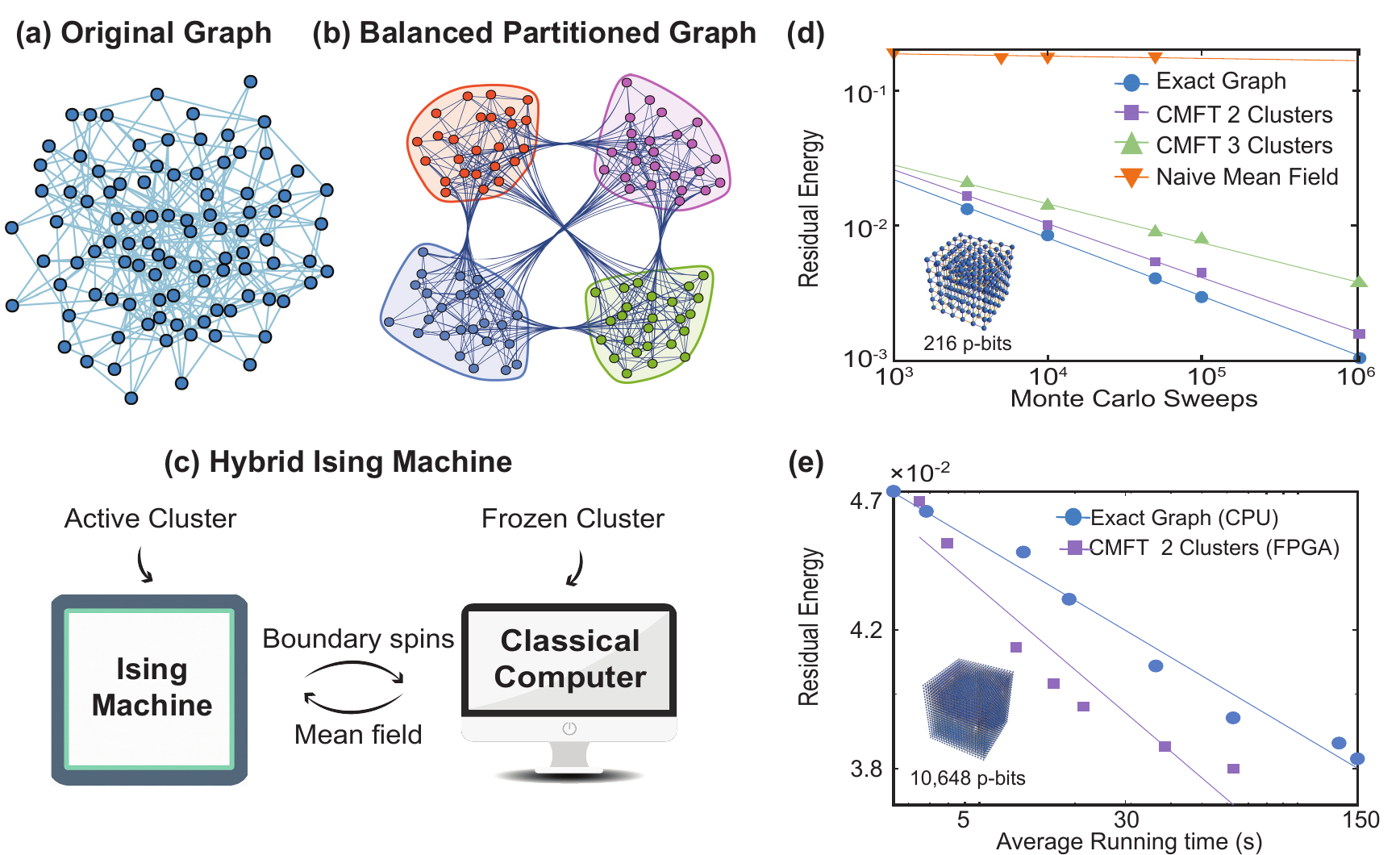}
    \caption{\textbf{Cluster mean-field theory for hardware 
capacity scaling.}
(a) An original interaction graph whose size exceeds 
the capacity of a single Ising-machine chip.
(b) Balanced partitioning of the graph into clusters sized 
to fit within hardware limits.
(c) Serial CMFT workflow: the active cluster is updated on 
the Ising machine while remaining clusters stay frozen. 
Boundary interactions are approximated by mean-field values 
computed on a classical processor.
(d) CPU results for $L=6$ (216 spins): residual energy 
versus Monte Carlo sweeps for different cluster counts 
(100 instances, 20 runs). Fewer clusters yield steeper 
power-law decay, where $\kappa$ denotes the log–log slope (exact graph: $\kappa=0.43$; two clusters: 
$\kappa=0.40$; three clusters: $\kappa=0.29$), while naive 
mean-field (each spin as its own cluster) is nearly flat 
($\kappa=0.02$).
(e) FPGA results for $L=22$ (10,648 spins): two-cluster 
CMFT enables problem sizes beyond single-FPGA capacity 
while reaching residual energies comparable to the 
unpartitioned CPU baseline at shorter wall-clock times 
(10 instances, 10 runs).}
    \label{fig:cmft}
\end{figure}

The annealing and tempering methods discussed so far assume that the relevant replicas or problem graphs can be mapped onto available hardware. A separate challenge arises when the interaction graph itself exceeds the number of physical spins or on-chip memory available on a single Ising-machine chip. We next turn to this capacity-scaling problem.

Cluster mean-field theory (CMFT) provides a hardware-agnostic route
to capacity scaling
\cite{yamamoto2009correlated,xing2003generalized}. Instead of mapping the
entire graph at once, the problem is decomposed into clusters sized
to fit the Ising machine's physical limits, with balanced partitions
identified heuristically via minimum-cut algorithms
(Figure~\ref{fig:cmft}b). Spins within the active cluster interact exactly on the hardware, while their couplings to external, frozen
spins are replaced by effective mean-field biases computed on a host
processor. By sequentially iterating
updates across all clusters, global information propagates through
the system, approximating exact MCMC sampling
(Figure~\ref{fig:cmft}c). Dynamic partitioning can further reduce boundary errors by ensuring that the same spins are not permanently assigned to cluster interfaces.

A key observation is that residual energies under CMFT still exhibit power-law scaling with Monte Carlo sweeps, with gracefully degrading
exponents as the number of clusters increases
(Figure~\ref{fig:cmft}d). As an example, for a 216-spin 3D spin glass, a two-cluster partition deviates only modestly from the unpartitioned
baseline, and even a three-cluster partition retains useful scaling behavior. Only in the extreme limit of one spin per cluster (naive mean field theory) does the optimization collapse entirely. This graceful degradation has a practical consequence: CMFT does not need to match exact solvers sweep-for-sweep. Because hardware acceleration can deliver orders of magnitude more sweeps per second, the modest
increase in sweeps required by the approximation is more than compensated by raw speed. Figure~\ref{fig:cmft}e demonstrates this
for a 10,648-spin problem ($L=22$) that exceeds single-FPGA capacity: the two-cluster CMFT implementation achieves a steeper
residual energy descent over wall-clock time than an unpartitioned CPU solver.

We note that CMFT also offers a natural algorithmic lens for understanding distributed Ising machines, where multiple chips
communicate through a shared bus or network. Whenever inter-chip communication is slower than intra-chip spin dynamics, the system is implicitly performing a form of cluster mean field: each chip treats the rest of the system through boundary values that lag behind its
own updates. This connection, and a detailed analysis of CMFT convergence and its FPGA implementation, will be presented elsewhere.

\noindent Similar scaling pressures appear in neural-network accelerators, where large models are partitioned across multiple devices using pipeline, tensor, or data parallelism \cite{huang2019gpipe,narayanan2019pipedream,qi2024zero,shoeybi2019megatron}. The analogy to CMFT is useful at the level of engineering constraints: both settings must divide a large problem into accelerator-sized pieces, balance local work against communication, and manage the cost of exchanging boundary information \cite{tanaka2021automatic}. CMFT approaches this problem from a statistical-physics perspective. A large Ising system is decomposed into clusters that are solved locally, while the influence of the external system is summarized by approximate mean fields at the boundaries. This gives a physics-based language for reasoning about approximate distributed computation, where the quality of the approximation depends on cluster size, correlation length, and update schedule. Thus, neural-network accelerator practice and CMFT point to related partitioning challenges from different directions, suggesting useful synergies between the two fields.

\section{Variational approaches for general Hamiltonians} \label{sec_variational}

The cluster mean-field approaches discussed previously demonstrate the power of a hybrid architecture, where a classical CPU and a probabilistic accelerator work in tandem to overcome hardware capacity limits. However, while CMFT scales the size of the problem, it still relies on traditional Monte Carlo methods such as simulated annealing or parallel tempering to update the spins within each cluster. These methods have fixed update rules, which limits their ability to adapt to problem structure or discover shorter non-equilibrium pathways through rugged landscapes. Variational algorithms offer a complementary route: instead of choosing a heuristic in advance, they optimize a parameterized stochastic process using samples returned by the hardware.

\subsection{Probabilistic approximate optimization algorithm}

\begin{figure}[t!]
    \centering
    \includegraphics[width=\linewidth]{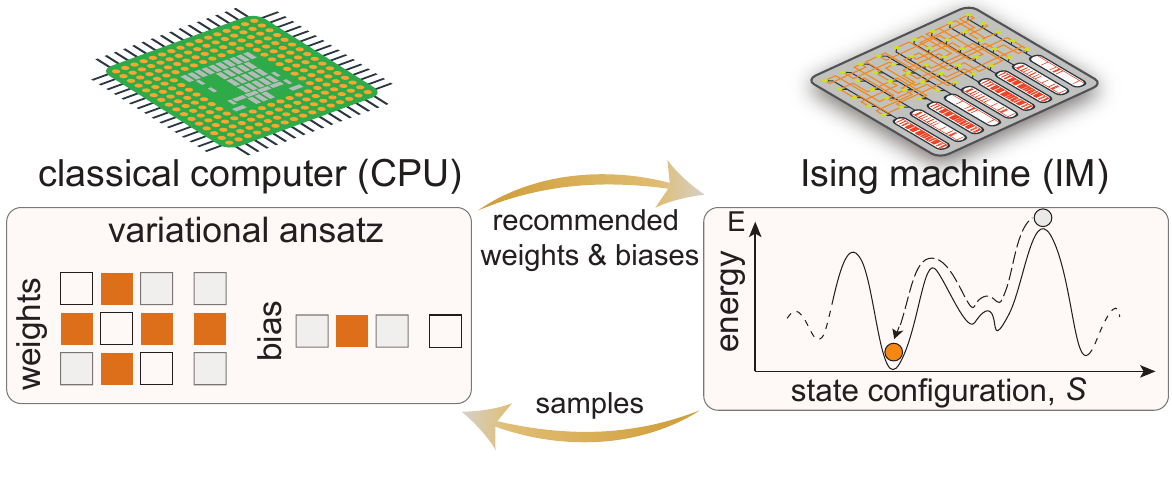}
   \caption{\textbf{Overview of PAOA.} A classical processor 
(left) evaluates a cost function using independent samples 
returned by an Ising machine or probabilistic sampler (right). 
The resulting objective is used to update variational parameters, 
such as effective couplings, fields, or annealing schedules, that 
define the stochastic dynamics for the next layer. Repeating this 
sample-and-update loop produces a learned non-equilibrium 
trajectory. In the illustrated ansatz, an $M \times M$ weight 
matrix and $M \times 1$ bias vector parameterize the sampling 
distribution over $N$ problem spins, where $M\geq N$.}
    \label{fig_PAOA}
\end{figure}

The Probabilistic Approximate Optimization Algorithm (PAOA) belongs to a broader family of classical variational dynamics inspired by QAOA \cite{farhi2022qaoa,zhou2020qaoa,QAOA_max_cut_farhi,sakai2025transferring,cerezo2021variational}. The original, groundbreaking PAOA proposal introduced classical probabilistic circuits built from low-dimensional stochastic matrices, showing that variational probabilistic processes can serve as meaningful classical counterparts to QAOA for combinatorial optimization \cite{Combes_2025}. In this view, the object being optimized is not a static energy function alone, but the stochastic transition process used to explore it.

This perspective has since been generalized to Markov-flow formulations over the full state space, with practical ansatze ranging from global temperature schedules to local fields, couplings, and multi-schedule annealing protocols \cite{PAOA_OPUS}. Simulated annealing appears as a restricted one-schedule limit of this broader variational family, while richer parameterizations can learn non-equilibrium trajectories adapted to the structure of a problem class. This flexibility also introduces an important caveat: highly expressive ansatze may overfit individual instances, so generalization requires either training over representative instance families or constraining the parameterization to lower-dimensional forms.

This trend is not limited to binary stochastic circuits. Recent work on the Variational Iterative Rotation Algorithm (VIRAL) develops a classical QAOA-like protocol based on kicked three-component spins \cite{morone2026variational}. Its performance on Sherrington--Kirkpatrick benchmarks suggests that part of the power of QAOA-like algorithms may come from their iterative variational structure, rather than from quantumness alone. This provides an independent example of the same broader idea: classical probabilistic or continuous-spin dynamics can form expressive variational families for optimization.

As illustrated in Figure~\ref{fig_PAOA}, PAOA alternates between an inner sampling loop and an outer optimization loop. The sampler generates independent samples from the current parameterized dynamics, while a classical processor updates the variational parameters to minimize a chosen objective. The inner loop can be accelerated by programmable Ising or probabilistic hardware; for example, an FPGA-based p-computer has been used to accelerate the sampling stage for 3D spin-glass problems \cite{PAOA_OPUS}. Other physical substrates may become relevant as the required stochastic or continuous-spin primitives mature.

Because the optimizer evaluates the actual samples produced by the hardware, variational schemes can also be made device-aware. In principle, reproducible nonidealities in the sampler can be incorporated into the objective being optimized, rather than treated only as errors to be calibrated away. Recent SPAD-based probabilistic hardware studies provide one example of this direction \cite{alsawidan2026}, although such device-specific implementations are best viewed as early demonstrations rather than the central motivation for PAOA. In practice, reproducible operating-point drifts are largely absorbed by retuning fields and couplings, while moderate device-to-device mismatch is tolerated as long as each unit's stochastic response remains monotonic.

The main open question is therefore not whether PAOA is a single killer application, but which restricted variational families, hardware substrates, and problem classes lead to reliable advantages. The search space is large: beyond global schedules, one can consider spatially inhomogeneous schedules, learned local temperatures, adaptive fields, or continuous-spin rotations. Related hand-designed strategies, such as inhomogeneous driving in quantum annealers and multiple non-equilibrium temperatures in MCMC, already suggest that spatially varying dynamics can be useful \cite{adame2020inhomogeneous,mohseni2021nonequilibrium}. PAOA and related classical variational dynamics provide a systematic way to learn such strategies rather than prescribe them by hand.

\subsection{Neural quantum states for quantum simulation}
Another prominent variational framework well suited to 
dedicated probabilistic hardware is neural quantum states  (NQS), in which many-body wavefunctions are parameterized  by neural networks and optimized using Monte Carlo sampling. 
Since their introduction by Carleo and Troyer 
\cite{carleo_solving_2017}, NQS have enabled accurate 
simulations of strongly correlated quantum systems and 
emerged as a powerful complement to quantum Monte Carlo 
and tensor-network methods  \cite{giammarco2019investigating,schmitt_quantum_2020,
netket3_2022,jvmc_2022,chen_empowering_2024,Moss2025leveraging}. The squared wavefunction amplitude $|\psi(s)|^2$ defines 
a positive distribution over spin configurations regardless  of whether the underlying amplitudes are real or complex,  so energy-based neural network ansätze map naturally onto  networks of p-bits, making sampling the central bottleneck 
to offload onto hardware.

Recent work has demonstrated this co-design by mapping 
sparse NQS onto FPGA-based probabilistic computers 
\cite{chowdhury2025probabilistic}. Enforcing locality in  the neural network connectivity reduces routing complexity  and enables simulations of two-dimensional quantum systems  with several thousand spins, where sampling becomes effectively negligible relative to the classical optimization steps. Crucially, the availability of a fast  dedicated sampler enables algorithmic advances such as 
conditional sampling, which allows deeper, more expressive  architectures to be trained without sacrificing hardware  compatibility. The same work also demonstrates that  multiple FPGA chips can be linked into a distributed  p-computer, with fast inter-chip communication ensuring  that boundary p-bits are exchanged exactly rather than  approximated, preserving correct Gibbs statistics across  partitions. This distributed scaling of Ising hardware 
is an active and broader research direction: as 
single-chip capacities plateau, multi-chip architectures  that maintain sampling fidelity will be essential for  tackling the larger problem sizes demanded by both optimization and quantum simulation.

\subsection{Positive probabilities and the sign problem}
\label{subsec:sign_problem}

Probabilistic Ising machines are most natural when the target object is a positive distribution, such as a Boltzmann distribution or the Born probability $|\psi(s)|^2$ used in variational Monte Carlo. In this setting, samples are ordinary positive events: the hardware draws states from a probability distribution, and observables are estimated from sample averages. Generic quantum circuits differ because their outputs are obtained by summing amplitudes over many computational paths, where each path may carry a positive, negative, or complex phase.

This distinction can be made precise by mapping quantum circuits to generalized Ising or Boltzmann-machine-like models with complex energies \cite{chowdhury2023emulating}. In such mappings, the probabilistic model can remain compact in circuit size, and sampling is still possible: one samples paths using the real part of the complex energy, then accounts for the imaginary part by attaching a phase factor to each sampled path. The cancellation happens after sampling, when these phase-weighted samples are accumulated to reconstruct the quantum amplitude. If many paths cancel, a large number of samples may contribute mainly to cancellation rather than to the final signal. This loss of useful signal is the Monte Carlo sign problem \cite{troyer2005computational}.

The sign problem is representation-dependent. A change of basis or a different decomposition can sometimes reduce cancellations, increase the useful post-cancellation signal, and make sampling far more effective \cite{hangleiter2020easing}. However, finding such a representation may itself be difficult, and no general efficient transformation is known that removes this obstruction for arbitrary non-stoquastic Hamiltonians or quantum circuits \cite{marvian2019computational}. This gives a useful way to state the boundary: probabilistic machines already operate in exponentially large state spaces, but with positive weights. At the very least, the sign problem identifies where genuine quantum advantage must come from: the controlled choreography of cancellations among amplitudes in the $2^N$-dimensional state space, or equivalently through $2^N \times 2^N$ transformations. When such cancellations are not essential, probabilistic sampling methods provide a natural classical competitor; when they are essential, they mark the regime where quantum mechanics is doing something that positive-probability sampling cannot generically reproduce. This distinction is especially important when evaluating QAOA-like algorithms against classical probabilistic variational methods.

\section{Synergies with generative AI} \label{sec_synergies_AI}
The algorithms discussed so far treat the Ising machine as a 
self-contained solver whose parameters are tuned by an outer 
loop. An emerging alternative is to pair the Ising machine 
with a generative model so that each compensates for the 
other's weakness: the generative model proposes global state 
updates that local MCMC moves would take exponentially long 
to discover, while the Ising machine verifies proposals 
through cheap energy evaluations that preserve exactness. 
The synergy can also run in the opposite direction: stochastic, 
additive primitives native to Ising hardware may accelerate 
inference in large-scale AI models themselves.

\subsection{Augmenting Ising machines with learned proposals}
\begin{figure}[t!]
    \centering
    \includegraphics[width=\linewidth]{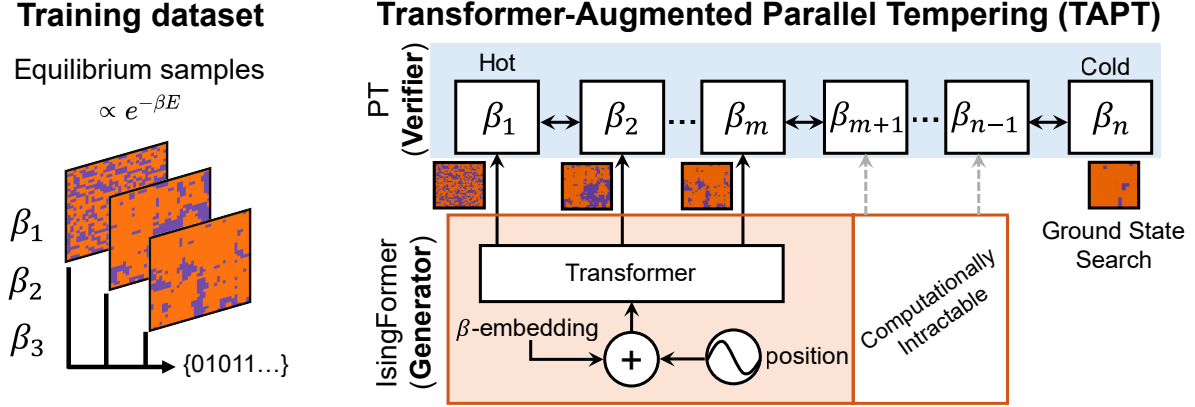}
       \caption{\textbf{Synergies between generative AI and Ising machines.} Left: Equilibrium samples collected from MCMC chains after burn-in serve as training data for a generative model. In this setup, a decoder-only Transformer (IsingFormer) is trained to produce equilibrium-like samples in fewer steps, effectively bypassing mixing times. Right: A Generator-Verifier scheme employing Transformer-Augmented Parallel Tempering (TAPT). Conditioned on the inverse temperature $\beta$, the IsingFormer acts as a generator to accelerate PT exploration via global moves along the temperature ladder. PT functions as a verifier, accepting or rejecting moves based on a Metropolis criterion to enforce detailed balance. The cold replicas remain computationally intractable for direct generation.}
    \label{fig:AI_Ising}
\end{figure}

Generative models have traditionally excelled in continuous domains where gradient-based methods are readily applicable \cite{jung2025roadmap}, but the discrete spins of the Ising model render the energy non-differentiable, making generative sampling fundamentally more challenging. Workarounds include spin relaxation, as illustrated by \cite{schuetz2022combinatorial}, where graph neural networks (GNNs) are trained on relaxed continuous versions of the binary variables, and the resulting continuous states are projected back to binary variables to obtain candidate solutions. Policy-gradient methods can also be applied to the Ising model. In these methods, trajectories or samples are collected during training, and a REINFORCE-type gradient estimator is used to optimize a chosen objective \cite{Williams1992}. For example, policy gradients can be used to learn efficient MCMC proposal policies \cite{bojesen2018policy}, or to train autoregressive models for estimating physical quantities such as free energy and magnetization \cite{wu2019solving,nicoli2020asymptotically}. A similar idea can be extended to diffusion models for combinatorial optimization \cite{sanokowski2024diffusion}, and more sophisticated policy-gradient formulations, such as proximal policy optimization (PPO), can be incorporated to find low-energy configurations of the Ising energy \cite{sanokowski2025scalable}. Other approaches include variational annealing methods, where generative models are trained under a temperature-annealing schedule to bias generated samples toward low-energy configurations \cite{hibat2021variational,ma2024message,zhang2025generative}. While these approaches act as powerful heuristics that can generalize across problem instances, they generally lack mathematical guarantees of reaching the true ground state.

A natural way to combine the global proposal power of generative models with the 
exactness of MCMC is through a generator-verifier scheme \cite{bunaiyan2025}. In this context, global proposals refer to non-local candidate moves that can accelerate equilibration, or help the MCMC sampler traverse distant regions of the energy landscape by exploiting learned heuristics. A Transformer trained on equilibrium samples (Figure~\ref{fig:AI_Ising}) proposes independent, non-local states that help the chain escape local minima. Because the model provides access to the exact configuration probability $q_\theta(s)$, a standard Metropolis--Hastings acceptance step can enforce detailed balance when exact sampling is needed \cite{nicoli2020asymptotically}. This idea was applied to sampling, where learned exchange moves augment PT replica exchanges and reduce the need for dense intermediate temperature rungs \cite{invernizzi2022skipping,zhang2026accelerated}. A recent ML-assisted Global Annealing scheme has also shown a clear advantage over tuned Simulated and Population Annealing on 3D Edwards--Anderson spin glasses \cite{delbono2026demonstrating}. In Transformer-Augmented Parallel Tempering (TAPT) \cite{bunaiyan2025}, the approach is extended to hard combinatorial optimization: an IsingFormer conditioned on inverse temperature $\beta$ generates global moves for multiple PT replicas (Figure~\ref{fig:AI_Ising}), accelerating the ground-state search on NP-hard instances such as 3D spin glasses and integer factorization compared to vanilla PT. The IsingFormer also generalizes across instances: trained on a fixed Ising multiplier circuit whose external biases specify the semiprime to factor, a single model transfers across an entire family of factorization instances without retraining, and on 2D Ising sampling it generalizes to unseen $\beta$ values including the critical region. 

A central open question for such schemes is \emph{generality} more broadly: most demonstrations train a separate model per instance, amortizing training cost only across runs of the same Hamiltonian. Extending TAPT-style transfer beyond invertible-logic encodings, for example by training over distributions of couplings $J_{ij}$ or by learning proposals conditioned on a ``master'' Hamiltonian that parameterizes a family of disorder realizations, is a natural and largely open direction for ML-assisted optimization.

\subsection{Probabilistic attention for the KV-cache memory wall}

Ising machines also share useful primitives with modern AI inference.
Recent work on SANTA \cite{lee2026_SANTA} gives a concrete example in
LLM autoregressive decoding. In standard attention, the weights over
cached tokens are given by
$A=\mathrm{softmax}(qK^T/\sqrt{d_k})$, which is mathematically a
Boltzmann distribution over token indices. Rather than streaming and
multiplying all $n_k$ value rows, SANTA samples $S \ll n_k$ indices
from this distribution and averages only the corresponding value
vectors. The resulting estimator is unbiased, admits variance reduction
through stratified or systematic sampling, and replaces the
post-softmax value multiplication with sampling, indexing, and addition.

This directly addresses the KV-cache memory-bandwidth bottleneck in
long-context decoding. SANTA kernels achieve up to a $1.5\times$
decode-step attention-kernel speedup over FlashInfer and FlashDecoding
at 32k-token contexts while matching baseline performance on reasoning
and long-context benchmarks \cite{lee2026_SANTA}. Unlike cache
compression or eviction, this approach does not permanently discard
past tokens; it reduces the number of value rows accessed at each
decoding step and is therefore complementary to KV-cache compression,
quantization, and grouped-query attention.

From the perspective of Ising hardware, the important point is that the
core operation is sampling from a Boltzmann-like distribution. This is
precisely the primitive implemented by p-bits and probabilistic Ising
samplers, and recent magnetic-tunnel-junction p-bit experiments have
demonstrated hardware softmax sampling \cite{softmax_pbit_2026}. Thus,
probabilistic/Ising-like algorithms can be useful not only for Ising
optimization problems, but also for reducing memory traffic in modern
AI inference.

\section{Outlook} \label{sec_outlook}

The algorithms surveyed in this Review share a common 
thread: each translates a principled probabilistic idea into a 
form that can be mapped onto physical Ising hardware, and 
in several cases the hardware itself has motivated entirely 
new algorithms. These examples -- from cluster mean-field 
decompositions and neural quantum states to 
Transformer-augmented tempering and stochastic sparse 
attention -- are necessarily a subset of a much larger 
algorithmic design space that will continue to grow as Ising 
platforms mature. The broader lesson is that the 
bidirectional flow between algorithms and hardware is the 
defining opportunity for the field going forward.

A recurring theme is the value of top-down, algorithm-first 
thinking. Simulated annealing, parallel tempering, and 
cluster mean-field methods all originated as software 
algorithms grounded in statistical mechanics, with 
well-understood convergence properties. When these methods 
are co-designed with hardware, the result is not merely a 
faster implementation but a qualitatively richer toolkit: 
two-dimensional parallel tempering natively encodes hard 
constraints that would otherwise require penalty terms, and 
cluster mean-field decompositions turn finite chip capacity 
into a feature rather than a limitation. By contrast, purely 
bottom-up approaches that rely on intrinsic device dynamics, 
while capable of impressive demonstrations, often lack the 
flexibility to adapt when problem structure changes. 
Continued investment in algorithms with provable guarantees 
provides the foundation on which hardware-specific 
optimizations can be layered.

Perhaps most striking is the emergence of algorithms that 
would not exist without the hardware. PAOA arose as a 
classical analog of QAOA specifically because probabilistic 
hardware made non-equilibrium driving schedules practical, 
and stochastic attention mechanisms repurpose the sampling 
and additive primitives native to Ising machines for 
accelerating large language model inference. These are 
likely early examples of a much broader trend: as Ising 
machines scale to larger sizes and higher speeds, they may 
open algorithmic possibilities that are currently impractical 
on conventional hardware.

As discussed in Section~\ref{subsec:sign_problem}, the sign problem gives a useful boundary for probabilistic Ising machines: positive-probability sampling can be extremely powerful, but genuine quantum advantage must ultimately rely on controlled interference among signed or complex amplitudes.

\section*{Acknowledgments}  
The authors acknowledge support from the Office of Naval Research (ONR), Multidisciplinary University Research Initiative (MURI) grant N000142312708, the National Science Foundation (NSF) CAREER Award under grant number CCF 2106260 and 2311295, the Army Research Laboratory under grant number W911NF-24-1-0228, the Institute of Energy Efficiency (IEE), and the Software Impact Grant Program.


\printbibliography

\end{document}